\documentclass[pdflatex,sn-basic]{sn-jnl}

\usepackage{graphicx}%
\usepackage{multirow}%
\usepackage{amsmath,amssymb,amsfonts}%
\usepackage{amsthm}%
\usepackage{mathrsfs}%
\usepackage[title]{appendix}%
\usepackage{xcolor}%
\usepackage{textcomp}%
\usepackage{manyfoot}%
\usepackage{booktabs}%
\usepackage{algorithm}%
\usepackage{algorithmicx}%
\usepackage{algpseudocode}%
\usepackage{listings}%
\usepackage{CJKutf8}%
\usepackage{url}%

\makeatletter
\g@addto@macro{\UrlBreaks}{\UrlOrds}
\makeatother

\newcommand{\maochar}{{\CJKfamily{bsmi}冇}}

\theoremstyle{thmstyleone}%
\theoremstyle{thmstyletwo}%

\theoremstyle{thmstylethree}%

\begin{document}
\begin{CJK*}{UTF8}{gbsn}

\title[Regional Slang Lexicon via Soft-Label Bidirectional Anchoring]{Constructing a Regional Slang Lexicon for Electricity Demands via Soft-Label Bidirectional Anchoring and Active Learning}

\author[1]{\fnm{Rong} \sur{Chang}}\email{rong\_chang007@126.com}

\author[1]{\fnm{Lijing} \sur{Shi}}\email{joshua.zang@outlook.com}

\author*[2,3]{\fnm{Zhaoxiang} \sur{Zang}}\email{zhaoxiang.zang@kust.edu.cn}

\affil[1]{\orgname{Yuxi Power Supply Bureau, Yunnan Power Grid Co., Ltd.}, \orgaddress{\city{Yuxi}, \postcode{653100}, \country{China}}}

\affil*[2]{\orgdiv{Faculty of Information Engineering and Automation}, \orgname{Kunming University of Science and Technology}, \orgaddress{\city{Kunming}, \postcode{650504}, \country{China}}}

\affil[3]{\orgdiv{Key Laboratory of Artificial Intelligence in Yunnan Province}, \orgname{Kunming University of Science and Technology}, \orgaddress{\city{Kunming}, \postcode{650504}, \country{China}}}

\abstract{%
Power-grid customer service increasingly relies on community social media channels, where customers describe outages, billing issues, and equipment faults in informal regional dialect. These short, noisy messages exhibit non-standard morphology, semantic drift, and out-of-vocabulary expressions that make manual lexicon construction costly and degrade conventional NLP tools. This paper proposes a weak-supervision framework for constructing a regional slang lexicon from curated dialect anchors and power-domain seed terms. The framework generates candidates via Pointwise Mutual Information with soft-label bidirectional anchoring, retaining dialect-anchored terms while recording power-domain relevance as separate metadata. Candidates are then scored by combining Sentence-BERT semantic similarity with a jieba-based wordhood measure of lexical fixedness. Uncertainty-aware active learning with rule-based and retrieval-augmented generation (RAG) LLM oracles is applied under three query strategies, and the resulting lexicon is independently evaluated by a higher-capability LLM calibrated against human expert annotations via Cohen's $\kappa$. Experiments on 11,177 real WeChat messages from Yuxi (Yunnan, China) achieve an evaluator-estimated dialect precision of DP=1.0 with zero false positives. A controlled ablation shows that soft-label layering retains 751\% more candidates than strict power-gating at identical precision, expanding the candidate pool from 41 to 349 terms. Human calibration ($n=46$) yields $\kappa_{\text{dialect}}=1.0$ and $\kappa_{\text{power}}=0.85$, indicating perfect agreement on dialect identification and strong agreement on power-domain relevance. The resulting lexicon, comprising 25--33 dialect terms per configuration of which 4--5 are additionally power-relevant, provides a preliminary resource for dialect-aware power-grid customer service NLP, though practical utility remains to be demonstrated through task-based evaluation.
}


\maketitle

\section{Introduction}\label{sec:introduction}

\subsection{Background and Motivation}\label{subsec:background_motivation}

The digital transformation of power-grid customer service under the Smart Grid 2.0 paradigm has changed how utilities interact with customers. In China, the State Grid Corporation and China Southern Power Grid have increasingly extended customer service from traditional voice hotlines such as 95598 to localized, community-based social media groups, especially WeChat communities organized by neighborhood, village, or transformer service area \citep{bib1,bib2}. This shift improves accessibility and response speed, but it also creates new challenges for automated intention recognition systems that were originally designed for formal and structured language.

The main difficulty is linguistic. In ethnically and linguistically diverse regions such as Yuxi City in central Yunnan Province, customers often report power outages, equipment faults, and billing inquiries using local dialectal expressions that differ from standard Mandarin in morphology, syntax, and semantics \citep{bib2,bib3}. The Yuxi dialect, a variety of Southwestern Mandarin, includes distinctive morphological markers (e.g., sentence-final particles such as 呢 (ne), 噶 (ga), 嘛 (ma), and 哈 (ha)), function words (e.g., the question marker 给有 (gei you, ``is there?'') and the intensifier 扎实 (zhashi, ``very'' or ``severe'')), and cases of semantic drift in which standard Chinese lexemes acquire domain-specific meanings \citep{bib4}. Table~\ref{tab:semantic_drift} illustrates representative cases of this phenomenon in power-grid contexts.

\begin{table}[t]
\caption{Semantic drift in Yuxi dialect terms within power-grid service contexts. Pinyin is provided for Chinese terms, and each example includes an English translation.}\label{tab:semantic_drift}
\centering\footnotesize
\begin{tabular}{p{0.16\textwidth}p{0.14\textwidth}p{0.19\textwidth}p{0.38\textwidth}}
\toprule
\textbf{Chinese term (pinyin)} & \textbf{Literal meaning} & \textbf{Power-grid meaning} & \textbf{Translated example} \\
\midrule
扎实 (zhashi)   & Sturdy, solid, firm  & Severe, very, intense           & ``雷打得扎实, 变压器都跳了'' (The thunder was intense; the transformer tripped). \\
鬼火 (guihuo)   & Phosphorescence, will-o'-the-wisp & Anger, frustration, annoyance & ``三天两头停电, 哪个不鬼火?'' (Outages every few days; who would not be frustrated?). \\
给有 (gei you)  & To give, to have      & Is there? Does it exist? (question marker) & ``供电局呢领导, 给有电了嘛?'' (Power bureau leaders, is there electricity yet?). \\
着炸 (zhao zha) & Affected + to explode & Damaged by electrical surge, shorted & ``表箱着炸掉了, 快点来看哈'' (The meter box was damaged by a surge; please come and check it). \\
老火 (laohuo)   & Old fire              & Severe, terrible, critical (intensifier) & ``线路老火, 快派人来修'' (The line fault is severe; please send someone to repair it). \\
\botrule
\end{tabular}
\end{table}

These expressions carry task-relevant semantic cues about fault severity (扎实 (zhashi) and 老火 (laohuo)), user emotional state (鬼火 (guihuo)), and request type (给有 (gei you)). For example, a customer service NLP system that interprets 线路老火 (xianlu laohuo, literally ``line old fire'') rather than ``severe line fault'' may misclassify complaint urgency and route the request incorrectly. Reliable handling of dialectal expressions is therefore important not only for model robustness, but also for maintaining service quality across linguistically diverse user populations \citep{bib5,bib6}.

\subsection{Technical Challenges}\label{subsec:technical_challenges}

Mining regional electricity slang from WeChat group messages presents three technical challenges that distinguish this setting from prior work on domain-specific NLP.

The first challenge is \textbf{data noise and sparsity}. Unlike formal corpora such as customer service hotline transcripts or technical maintenance logs, WeChat group messages contain heterogeneous noise that can be grouped into three categories. \emph{Structural noise} originates from the WeChat platform itself, including system-generated messages (e.g., ``拍了拍'' (pai le pai, profile tap notification) and ``邀请加入了群聊'' (invited a user to join the group chat)), XML-formatted metadata (\texttt{<msg>...</msg>} tags), \texttt{wxid\_} user identifiers embedded in export artifacts, and application-specific formatting that survives CSV export \citep{bib7}. \emph{Irrelevant noise} includes cross-domain content that appears in power service groups, such as commercial advertisements, community administrative notices, public health announcements, and casual social chatter unrelated to electricity service \citep{bib8}. \emph{Visual noise} arises from rapid mobile input, including typographical errors, homophone substitutions, and bracket-delimited emoji placeholders such as \texttt{[捂脸]} ([facepalm]) \citep{bib7}. In aggregate, structural noise accounts for over 30\% of raw message volume, and the overall signal-to-noise ratio---the fraction of messages containing genuine electricity-related content---falls below 5\% \citep{bib9}.

The second challenge is \textbf{term ambiguity in short texts}, a well-documented phenomenon in short-text stream clustering \citep{bib10,bib11}. A single WeChat message typically contains 10--30 Chinese characters, leaving limited context for disambiguation. The Yuxi dialect intensifier 扎实 (zhashi), which means ``severe,'' ``very,'' or ``intense'' in power-grid complaints, retains the standard Mandarin meaning of ``sturdy'' or ``solid'' in other domains. A message such as ``扎实呢停电'' (zhashi ne tingdian, roughly ``the outage is severe'') therefore requires the system to recognize both the dialect particle 呢 (ne) and the context-dependent meaning of 扎实, a combination that is difficult to resolve without domain-specific anchoring \citep{bib12}.

The third challenge is the \textbf{out-of-vocabulary (OOV) and few-shot dilemma}. General-purpose pre-trained language models, including Chinese RoBERTa variants and multilingual BERT, are trained mainly on standard Mandarin corpora from news, encyclopedias, and formal publications \citep{bib13,bib14}. Dialect-specific morphology, character substitutions, and regional expressions are often poorly represented in these corpora, which can degrade zero-shot transfer performance \citep{bib15}. Although fine-tuning on domain-specific annotated data could mitigate this problem, constructing a sufficiently large labeled corpus for a low-resource dialect is costly \citep{bib16,bib17}. As a result, standard NLP pipelines deployed in dialect-heavy regions may perform substantially worse than they do on standard-language benchmarks \citep{bib18}.

\subsection{Limitations of Existing Approaches}\label{subsec:limitations_existing}

Existing approaches to power-grid NLP have primarily focused on structured, formal-language settings. Prior work has established foundations for named entity recognition (NER) and dispatch classification using annotated transcripts of 95598 hotline calls, which are characterized by a professional register, standard terminology, and relatively clean transcription. Multi-task learning frameworks have been proposed for few-shot power dispatch NER \citep{bib19}, and deep active learning algorithms have been applied to Chinese power text classification to balance annotation cost and model accuracy \citep{bib20}. These methods generally assume that input text follows formal register conventions and that vocabulary is drawn from standard technical terminology with manageable OOV rates. These assumptions are problematic for WeChat group messages, where dialect-inflected expressions create vocabulary coverage gaps on the order of 15--25\% \citep{bib7}.

Beyond the power domain, computational approaches to dialect processing have focused mainly on dialect identification, namely classifying text by regional origin, rather than on the constructive problem of building a dialect lexicon from unlabeled text \citep{bib21,bib22}. Studies in Arabic dialectology \citep{bib23}, Swiss German processing, and Chinese dialect identification \citep{bib24} show that statistical features can detect regional variation, but these methods typically assume the availability of reference resources and do not address how to bootstrap a lexicon from noisy, unlabeled data. Weak supervision techniques \citep{bib25,bib26} and data programming frameworks \citep{bib25,bib27} provide foundations for generating training signals without full manual annotation, but their application to dialect lexicon construction remains insufficiently explored.

\subsection{Proposed Approach and Contributions}\label{subsec:approach_contributions}

We formulate regional slang lexicon construction as a noisy short-text semantic expansion problem and propose a four-stage weak-supervision framework that uses a small set of curated dialect anchors and power-domain seed terms rather than a fully labeled training corpus. The key idea is that verified dialect anchor words, i.e., morphemes known to be characteristic of the target dialect, can serve as statistical anchors for discovering additional dialect expressions through Pointwise Mutual Information (PMI) \citep{bib28}. This anchor-driven design is combined with a soft-label bidirectional anchoring mechanism that records power-domain relevance without using it as an early filtering condition. In this way, the framework separates noise elimination from relevance assessment, two tasks that are often conflated in hard-gating approaches.

The pipeline proceeds in four stages. Stage 1 extracts bigram candidates using PMI anchored on 16 Yuxi dialect markers, with bidirectional anchoring, soft-label power-domain layering, and a four-condition fragment filter used to remove grammatical constructions while preserving lexical compounds. Stage 2 classifies candidates using a dual-channel score that combines Sentence-BERT semantic similarity to 65 power-domain seed terms with a jieba-based wordhood score that measures lexical fixedness through segmentation resistance \citep{bib24,bib29}. Stage 3 performs uncertainty-aware active learning with three query strategies (Margin, TD-CAL, Random) \citep{bib30,bib31,bib32} and two oracle types: rule-based morphological matching and RAG-augmented LLM assessment \citep{bib33} using DeepSeek V4 Flash \citep{bib34}. Stage 4 evaluates the constructed lexicon with a separate, higher-capability LLM (DeepSeek V4 Pro) calibrated against human expert annotations via Cohen's $\kappa$ \citep{bib35}, while retaining support for cross-model evaluation using a different model family.

We evaluate the framework on 11,177 real WeChat messages from Yuxi power-grid customer service groups and use a controlled ablation study to isolate the contribution of soft-label layering. The study makes four contributions. First, it introduces a bidirectional anchoring framework with soft-label power-domain layering that reaches an LLM-evaluator-estimated dialect precision of DP=1.0 while increasing candidate coverage by 751\% over strict power-gating. Second, it presents a dual-track LLM evaluation system with an independent evaluator model calibrated against human expert annotations, yielding degenerate perfect agreement for dialect identification ($\kappa_{\text{dialect}}=1.0$) and strong agreement for power-domain relevance ($\kappa_{\text{power}}=0.85$, $n=46$), while noting that cross-model validation remains future work. Third, it compares two oracle types across three active learning query strategies and shows that oracle quality affects strategy differentiation: under the high-acceptance LLM oracle ($\sim$89\%), all configurations converge to 4 power-dialect terms, whereas modest differentiation appears under the more selective rule-based oracle. Fourth, the ablation shows that soft-label layering preserves many dialect bigrams discarded by hard gating while maintaining the same dialect precision, suggesting a practical design principle for weak-supervision pipelines in low-resource settings.

\section{Related Work}\label{sec:related_work}

\subsection{Natural Language Processing in Power Systems}\label{subsec:nlp_power_systems}

NLP has been increasingly applied to power systems engineering and customer service as utilities accumulate unstructured text from maintenance logs, fault reports, hotline transcripts, and online service channels. Early work on message understanding and information extraction provides a methodological background for structured extraction from operational text \citep{bib36}. More recent work has introduced annotated hotline transcript datasets with entity and dispatch category labels, enabling baseline comparisons for architectures such as BiLSTM-CRF and BERT-based classifiers.

Several extensions have addressed limitations of these baselines. Multi-task learning frameworks combining NER with relation extraction have improved nested entity recognition in power dispatch contexts \citep{bib19}, while domain-adaptive pre-training can reduce vocabulary mismatch between general-domain language models and domain-specific terminology \citep{bib37}. In fault diagnosis, deep active learning has reduced annotation requirements for power equipment fault classification while maintaining accuracy above 90\% \citep{bib20}. Knowledge-enhanced methods that incorporate power-grid topology into text encoders further support complex fault reasoning tasks.

However, most power-grid NLP studies assume text that is relatively clean, formally structured, and drawn from standard technical vocabulary. Hotline transcripts, maintenance logs, and dispatch records are typically produced in professional settings with controlled terminology and standardized reporting formats. Community chat logs differ in both form and content: they contain non-standard syntax, regional dialect, code-switching, emoji usage, and substantial topic drift \citep{bib8}. This difference is qualitative rather than only quantitative, because informal community messages introduce different noise distributions and different model failure modes.

\subsection{Short Text Processing and Statistical Association Measures}\label{subsec:short_text_pmi}

The processing of short and sparse documents has been widely studied in information retrieval and text mining. Short texts, often defined as documents with fewer than 50 words, provide limited within-document co-occurrence evidence and therefore degrade the performance of standard topic models and clustering methods \citep{bib10}. The Topic-Enhanced Dirichlet Model (TEDM) addresses this problem by incorporating auxiliary word embeddings into topic inference, showing that external semantic knowledge can compensate for sparse local evidence \citep{bib11}. The Online Semantic-Enhanced Graphical Model (OSGM) extends this idea to streaming scenarios by dynamically maintaining topic distributions as new short documents arrive \citep{bib10}. These studies motivate the use of external semantic information in short-text analysis, which in our setting is provided by power-domain seed terms and dialect anchors.

Statistical association measures are central to unsupervised term extraction and collocation discovery. Since the work of Church and Hanks on word association norms \cite{bib28}, Pointwise Mutual Information (PMI) has been widely used to measure statistical dependence between two words by comparing observed co-occurrence with the expected frequency under independence. In computational lexicography, PMI helps identify multi-word expressions, technical terms, and domain-specific collocations that may be missed by frequency-based methods alone \citep{bib38}. Its normalized variant, NPMI, bounds association scores to $[-1, 1]$, which facilitates cross-corpus threshold selection and improves robustness to corpus size variation \citep{bib39}.

In Chinese NLP, PMI-based methods face an additional difficulty because word boundaries are not explicitly marked. Segmentation is therefore required before association scores can be computed, and segmentation errors, especially on OOV terms and dialect expressions, can propagate into PMI estimation \citep{bib40}. The jieba tokenizer provides a practical balance between speed and accuracy through its prefix dictionary and Hidden Markov Model-based OOV handling \citep{bib41}. Our pipeline builds on jieba but explicitly registers dialect anchor words and power-domain seeds before tokenization to reduce fragmentation of dialect expressions.

\subsection{Active Learning for Low-Resource Natural Language Processing}\label{subsec:active_learning}

Active learning (AL) refers to a family of methods that selectively query informative unlabeled instances for annotation, with the goal of improving models under limited labeling budgets \citep{bib30}. In the standard pool-based setting, a model is trained on a small labeled set, a query strategy selects instances from a larger unlabeled pool for oracle annotation, and the newly labeled instances are added to the training set. This cycle continues until the annotation budget is exhausted or model performance saturates.

The choice of query strategy affects AL efficiency. Uncertainty sampling selects instances for which the current model is least confident, using measures such as entropy, least confidence, or margin \citep{bib31}. Margin sampling selects instances with the smallest difference between the top two predicted class probabilities and is often effective for binary classification with a well-defined decision boundary \citep{bib31,bib42}. Density-weighted methods add a representativeness criterion, preferring instances that are both uncertain and located in dense regions of the feature space \citep{bib32}. Token-Density Contrastive Active Learning (TD-CAL) follows this principle by combining a margin-based uncertainty term with a density term based on token frequency and document entropy. In our setting, however, the standard TF-IDF-based density signal is nearly constant for rare dialect bigrams, all of which appear 2--3 times in the corpus. We therefore use PMI as a more discriminative density proxy, with values ranging from 2.4 to 11.8 for attested bigrams.

AL has been studied across NLP tasks including text classification \citep{bib42}, named entity recognition \citep{bib43}, relation extraction, and machine translation \citep{bib44}. Deep active learning combines AL with pre-trained representations, often coupling the query strategy with fine-tuning of Transformer-based models \citep{bib45}. In the power domain, Deng \emph{et al.} \cite{bib20} showed that uncertainty-based AL with BERT embeddings reduced annotation requirements by approximately 60\% for Chinese power text classification while maintaining accuracy above 90\%. Across these studies, oracle quality and consistency are important: noisy or inconsistent oracles reduce the informativeness of selected instances and weaken the benefit of active sampling \citep{bib30}. Our experiments examine this issue in an LLM-oracle setting, where the RAG-LLM oracle's high acceptance consistency ($\sim$89\%) reduces differentiation among query strategies.

\subsection{Computational Dialectology and Lexicon Construction}\label{subsec:dialectology_lexicon}

Computational studies of dialect variation have expanded over the past decade with the availability of geo-located social media data and advances in representation learning \citep{bib21}. The dominant task has been dialect identification: given a text sample, determine which regional variety of a language it represents. Shared tasks such as the DSL (Discriminating between Similar Languages) challenge \citep{bib23} and the VarDial evaluation campaign \citep{bib46} have advanced methods based on character n-gram features, word embeddings, and neural architectures for dialect classification across Arabic, Spanish, Portuguese, and Chinese varieties.

Dialect feature detection is a more fine-grained task that identifies the morphological, lexical, or syntactic markers associated with a dialect. Work on Swiss German dialects has shown that neural models can detect regional morphological patterns from raw text, while studies in Arabic dialectology \citep{bib47} and Chinese dialect variation indicate that character-, word-, and syntax-level features provide complementary signals for regional classification. However, these methods usually operate in supervised or semi-supervised settings that require annotated dialect data, which is limited in the setting considered here.

The constructive problem of building a dialect lexicon from unlabeled text has received less attention. Traditional dialect lexicon construction often relies on manual curation by domain experts, as illustrated by regional vocabulary compilations such as the Yuxi Dialect Vocabulary Compilation \citep{bib4}. This approach is precise but difficult to scale across dialects and domains. Distributional methods that cluster words using contextual embeddings have been applied to general-domain lexicon induction \citep{bib48}, but dialect terms are often too sparse for such methods to work reliably without additional constraints \citep{bib49}. Weak supervision methods, including data programming \citep{bib25,bib27} and Snorkel-based frameworks \citep{bib50}, provide a basis for combining multiple noisy labeling functions. Their use for dialect lexicon construction, however, requires task-specific labeling functions, which motivates the bidirectional anchoring approach proposed in this study.

Semantic drift, in which words change meaning across time periods, communities, or domains, has been studied using distributional semantics. Hamilton \emph{et al.} \cite{bib51} showed that word embedding spaces can track systematic historical semantic change and reveal patterns in the rate and direction of meaning shift. Our setting involves a synchronous form of semantic drift: the same character sequence (e.g., 扎实 (zhashi), ``sturdy'' in standard Mandarin but ``severe'' in Yuxi power-grid complaints) can carry different meanings in standard Chinese and in Yuxi power-grid service contexts. The SBERT-based semantic classification stage is designed to capture this domain-specific shift by measuring proximity to power-domain seed embeddings.

\subsection{Large Language Models for Evaluation and Annotation}\label{subsec:llm_evaluation_annotation}

Large language models (LLMs) are increasingly used for automated evaluation and annotation in NLP research \citep{bib52}. They have been applied to text quality assessment \citep{bib53}, factuality verification \citep{bib54}, and semantic similarity judgment \citep{bib55}, often reaching agreement levels close to human annotators. LLMs have also been explored as oracles in active learning pipelines, where they can match or exceed rule-based oracles on some semantic classification tasks. At the same time, their use requires caution because of consistency limitations, hallucination risk, and annotation cost.

A central methodological concern in LLM-based evaluation is circularity when the same model family is used for both selection and evaluation \citep{bib52,bib56}. Recent studies have also documented judge bias and calibration challenges in LLM-based evaluation frameworks \citep{bib57}. Even when different capability tiers are used, as in our setup with DeepSeek V4 Flash for selection and DeepSeek V4 Pro for evaluation, shared training data and model-family biases may still affect the evaluation. We address this concern in two ways. First, we calibrate the LLM evaluator against independent human annotations using Cohen's $\kappa$, which provides a reliability measure that does not rely on model-internal confidence scores \citep{bib35}. Second, the codebase supports replacing the evaluator with a different model family (e.g., GLM-4-Plus) for cross-model validation, although this robustness check has not yet been empirically executed.

Cohen's $\kappa$, introduced by Cohen \cite{bib35} and widely adopted in the behavioral sciences and NLP, measures inter-rater agreement for categorical labels while correcting for chance agreement. Values above 0.6 are conventionally interpreted as substantial agreement, and values above 0.8 as near-perfect agreement \citep{bib58}. Applying $\kappa$ to LLM evaluation calibration is a recent practice \citep{bib59} that supports more transparent reliability assessment of LLM-generated annotations in research settings.

\section{Methodology}\label{sec:methodology}

\subsection{Problem Formulation}\label{subsec:problem_formulation}

Let the streaming WeChat corpus be $D = \{d_1, d_2, \dots, d_n\}$, where each $d_i$ is a raw message string containing text, metadata, and noise artifacts. A preprocessing filter $F(\cdot)$ maps each raw message to a cleaned string, producing the preprocessed corpus $D'$. The goal is to construct a domain-specific lexicon $L = \{(w_1, c_1), \dots, (w_k, c_k)\}$, where $w_j$ is a dialect expression and $c_j \in \mathcal{C}$ denotes its semantic category. The category set $\mathcal{C}$ includes OUTAGE, REPAIR, EQUIPMENT, FAULT, BILLING, and OTHER, corresponding to the main dimensions of power-grid customer service interactions. In the current pipeline, these categories serve two internal purposes: organizing the 65 power-domain seed words into thematic groups (e.g., 停电 (tingdian, power outage) and 没电 (meidian, no electricity) are both OUTAGE seeds) and assigning each candidate a provisional category label based on its nearest seed in the SBERT embedding space. The evaluation, however, focuses only on two binary dimensions, \texttt{has\_dialect\_features} and \texttt{power\_domain\_relevance}, and does not assess category-label correctness. Category-level classification accuracy is therefore not measured, and the final dual-lexicon outputs ($L_d$ and $L_{pd}$) do not include category labels in their primary quality metrics. We retain the category structure as scaffolding for future downstream integration while making this scope limitation explicit.

\begin{figure}[t]
\centering
\includegraphics[width=\textwidth]{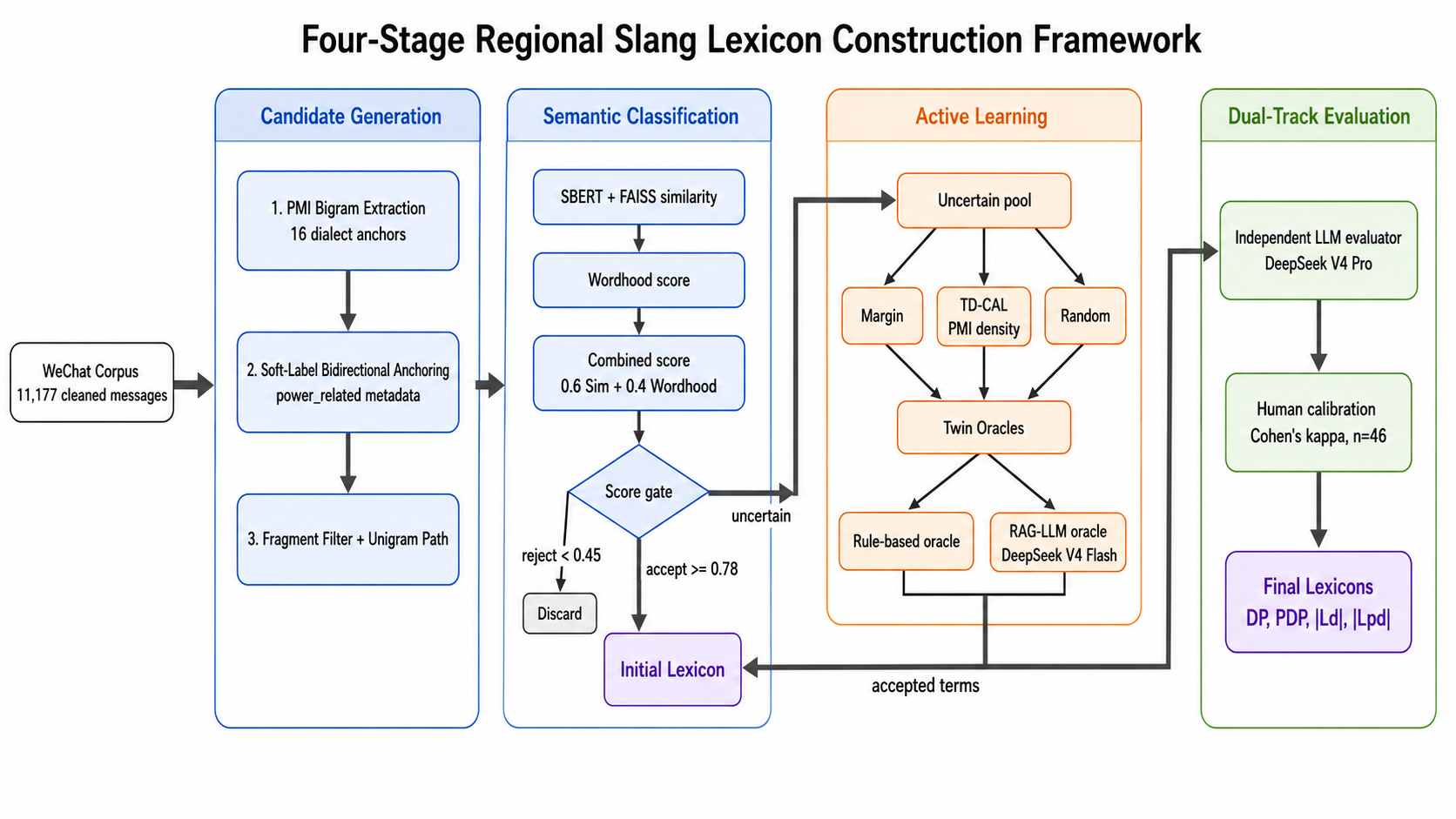}
\caption{Overview of the four-stage pipeline.}\label{fig:pipeline_arch}
\end{figure}

Figure~\ref{fig:pipeline_arch} summarizes the overall pipeline. The four stages are described in the following subsections.

\subsection{Stage 1: Candidate Generation with Bidirectional Anchoring}\label{subsec:stage1_candidate_generation}

\begin{figure}[t]
\centering
\includegraphics[width=\textwidth]{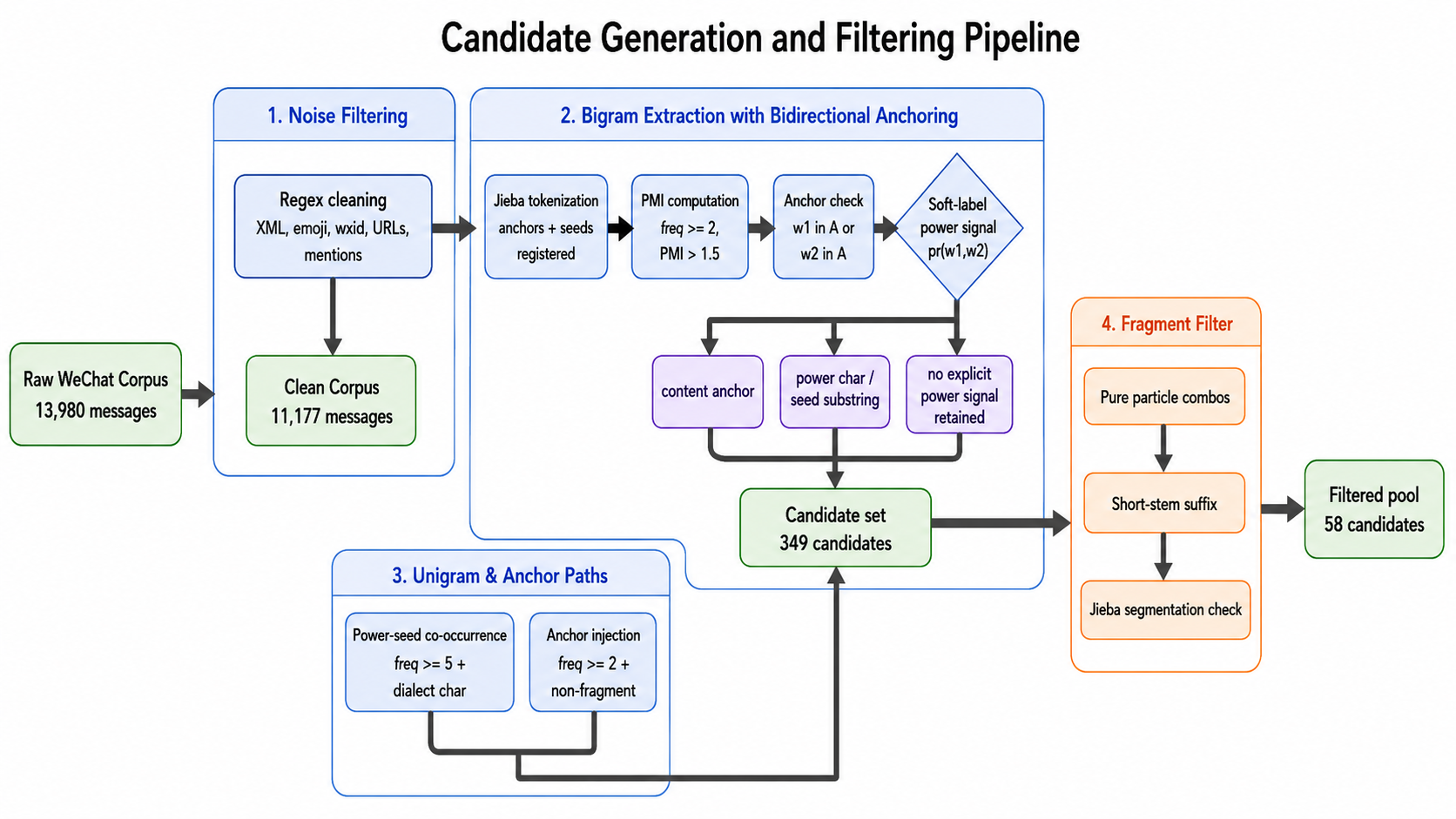}
\caption{Candidate generation pipeline: from raw corpus to filtered candidate pool.}\label{fig:candidate_generation}
\end{figure}

Figure~\ref{fig:candidate_generation} shows the four-step candidate generation process from the raw corpus to the filtered candidate pool.

\subsubsection{Adaptive Noise Filtering}\label{subsubsec:noise_filtering}

Raw WeChat messages contain metadata and platform artifacts that must be removed before linguistic analysis. We apply an ordered set of regular-expression transformations based on noise patterns observed in the corpus. XML-formatted system messages are removed by matching \texttt{<msg>...</msg>} patterns with recursive nesting support. Platform interaction notifications, including profile tap messages (``拍了拍'' (pai le pai)), group invitation messages (``邀请加入了群聊'' (invited a user to join the group chat)), and message recall notifications (``撤回了一条消息'' (recalled a message)), are removed by matching known templates. User identifiers in the \texttt{wxid\_} format are extracted and discarded. Application-specific bracket-delimited emoji placeholders (\texttt{[捂脸]} ([facepalm]), \texttt{[呲牙]} ([grinning face]), etc.), URLs, and @-mentions are also stripped.

After noise removal, messages are filtered to retain only Chinese characters (Unicode range U+4E00--U+9FA5), ASCII letters and digits, and a restricted set of punctuation marks (commas, periods, question marks, and exclamation marks). Messages with fewer than two Chinese characters after filtering are discarded as uninformative. From 13,980 raw messages collected over a one-year period, the preprocessing pipeline retained 11,177 cleaned messages, corresponding to a retention rate of approximately 79.9\%.

\subsubsection{Anchor-Driven Bigram Extraction}\label{subsubsec:anchor_bigram}

Candidate generation uses a curated set of Yuxi dialect anchor markers as statistical pivots for discovering associated expressions. The anchor set $\mathcal{A}$ contains 16 high-frequency Yuxi dialect morphemes selected from the \emph{Yuxi Dialect Vocabulary Compilation} and reviewed by domain experts: 呢 (ne), 挨 (ai), 咋个 (zage, ``how/what''), 给有 (gei you, ``is there?''), 是呢 (shi ne, ``is it?''), 扎实 (zhashi, ``very/severe''), 鬼火 (guihuo, ``frustrated''), 噶 (ga), 木 (mu), \maochar{} (mao, ``not have''), 咋整 (za zheng, ``what to do''), 着 (zhao, dialect passive/affected marker), 嘛 (ma), 哈 (ha), 得行 (de xing, ``workable/acceptable''), and 老火 (laohuo, ``severe/critical''). These anchors are divided into two functional groups. \emph{Function word anchors} (呢 (ne), 噶 (ga), 嘛 (ma), 哈 (ha), 咋个 (zage), 给有 (gei you), 着 (zhao), 是呢 (shi ne), 咋整 (za zheng), 挨 (ai), 木 (mu), and \maochar{} (mao)) act as syntactic markers of dialect-specific grammatical constructions. \emph{Content word anchors} (扎实 (zhashi), 鬼火 (guihuo), 老火 (laohuo), and 得行 (de xing)) carry independent semantic meaning and can combine with domain terms to form dialect compounds.

All anchors and 65 power-domain seed words spanning the five service categories are registered with the jieba Chinese tokenizer so that they are preserved as atomic tokens during segmentation. In addition, 839 reference dialect terms, defined as entries of length $\geq 2$ from the Yuxi Dialect Vocabulary Compilation, are registered with jieba to reduce fragmentation of known dialect expressions. For example, this step keeps 扎实 (zhashi, ``very/severe'') as one token rather than splitting it into 扎 + 实, two individual characters without the dialect-specific compound meaning. This registration is used only for tokenization calibration and does not constitute label leakage: jieba is told which character sequences can be treated as words, not which candidates are correct lexicon entries. The corpus is then tokenized with these registrations active. For each pair of adjacent tokens $(w_1, w_2)$ whose co-occurrence frequency $f(w_1, w_2) \geq 2$, the Pointwise Mutual Information is computed:

\begin{equation}
\mathrm{PMI}(w_1,w_2)=\log_2 \frac{P(w_1,w_2)}{P(w_1)P(w_2)}
=\log_2 \frac{f(w_1,w_2)\cdot T}{f(w_1)\cdot f(w_2)} .
\label{eq:pmi}
\end{equation}

Here, $f(\cdot)$ denotes token frequency in the jieba-segmented corpus and $T$ is the total number of tokens. The merged form $w_1w_2$ is retained as a candidate if at least one component belongs to $\mathcal{A}$, the PMI exceeds 1.5, and the merged form contains at least two Chinese characters. The PMI threshold of 1.5 was selected to balance recall and precision, with reference to prior work on collocation extraction using PMI \citep{bib28}. Normalized PMI is retained only as a diagnostic value and for comparison with earlier pipeline variants:

\begin{equation}
\mathrm{NPMI}(w_1,w_2)=
\frac{\mathrm{PMI}(w_1,w_2)}{-\log_2 P(w_1,w_2)} .
\label{eq:npmi}
\end{equation}

In the final experiments, NPMI is not used for candidate filtering; all six configurations use the raw PMI threshold described above. Beyond the original 16 anchors, dialect content-word anchors can be added from the reference vocabulary when they appear in the corpus. In the current corpus, this procedure added one content-word anchor.

\subsubsection{Soft-Label Bidirectional Anchoring}\label{subsubsec:soft_label}

An important design decision is how to handle dialect bigrams that do not contain explicit power-domain lexical content. A hard power-gating approach would discard any candidate whose component words do not contain electricity-domain characters or match power seed terms. This would remove dialect conversational markers such as 咋个整 (zage zheng, ``what to do'') and 那咋整 (na zage zheng, ``then what to do''), which occur in outage complaint threads and may carry indirect service-domain value. We therefore use a soft-label approach: every dialect-anchored bigram that passes the PMI threshold is retained in the candidate pool, and two independent metadata tags are used to separate dialect anchoring from power-domain relevance. The \texttt{power\_related} tag captures power-domain lexical content through character-level and seed-term matching:

\begin{equation}
\mathrm{pr}(w_1,w_2)=\mathbf{1}\bigl[
p_c(w_1)\vee p_c(w_2)\vee p_c(w_1w_2)
\bigr].
\label{eq:power_related}
\end{equation}

Here, $p_c(w)$ (\texttt{power\_chars}) is true when any character of $w$ belongs to a set of 21 characters commonly occurring in electricity-domain vocabulary (电, 线, 路, 闸, 表, 费, 火, 跳, 停, 修, 炸, 漏, 烧, 压, 杆, 变, 容, 开, 关, 供, 配; representative meanings include electricity, line, meter, outage, repair, and transformer-related concepts), or when $w$ contains a substring from a supplementary set of 65 short power-domain seed terms (2--3 character subsets). Substring matching captures terms such as 负荷 (fuhe, load) and 来电 (laidian, power restoration or incoming electricity) that express power-domain concepts but do not contain the explicit power characters.

\begin{figure}[t]
\centering
\includegraphics[width=\textwidth]{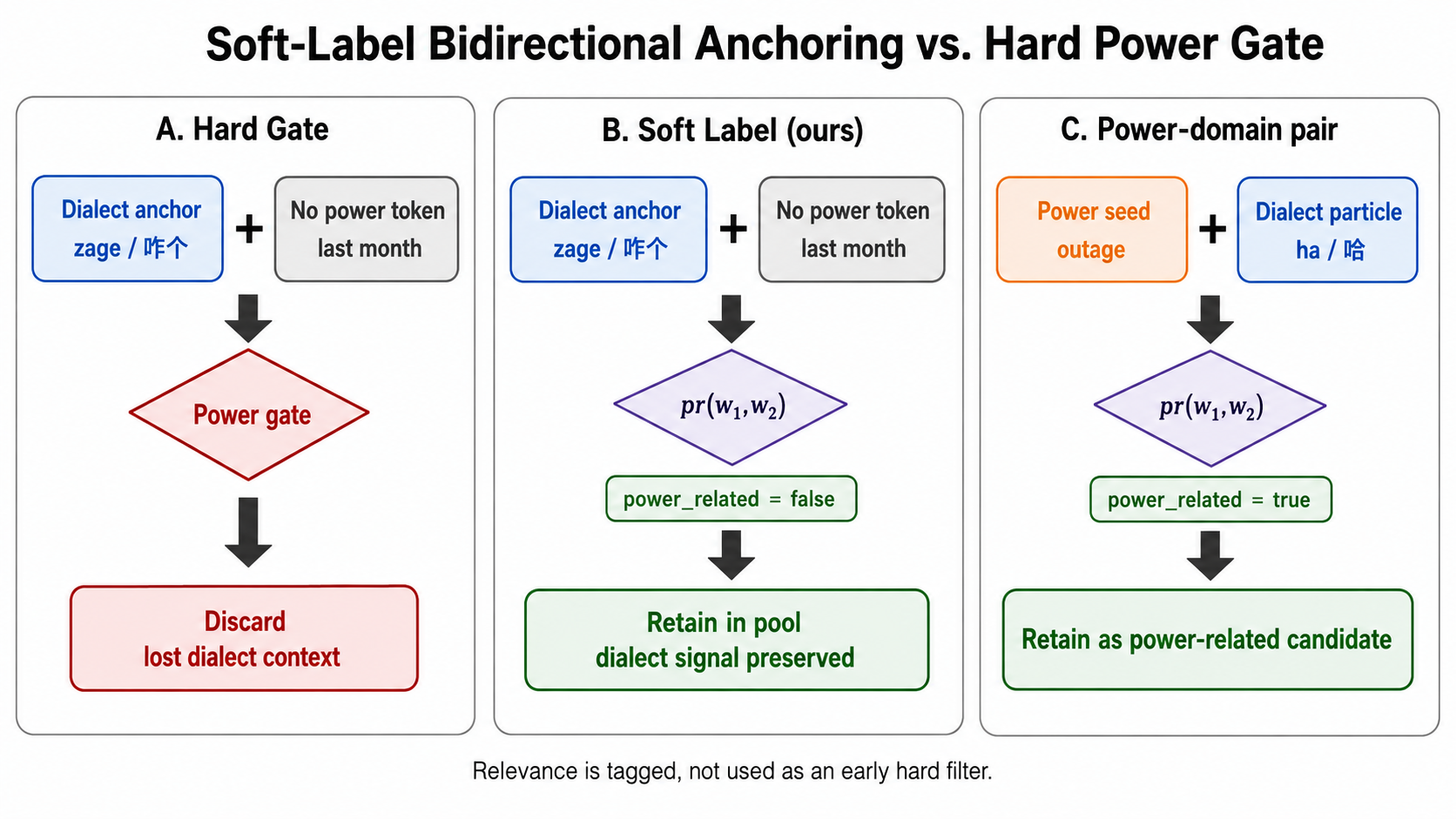}
\caption{Bidirectional anchoring: hard-gate versus soft-label comparison.}\label{fig:bidirectional_anchoring}
\end{figure}

Separately, a \texttt{has\_dialect\_anchor} tag records whether at least one component of the bigram is a content-word anchor ($\in \mathcal{A}\setminus\mathcal{A}_{\text{func}}$). Content-word anchors such as 扎实 (zhashi, ``very/severe''), 鬼火 (guihuo, ``frustrated''), 老火 (laohuo, ``severe/critical''), and 得行 (de xing, ``workable/acceptable'') carry independent semantic meaning and can combine with domain terms to form dialect compounds (e.g., 线路老火 (xianlu laohuo, ``severe line fault'')). This tag is stored for ablation analysis but does not affect the \texttt{power\_related} computation; the two signals are kept independent. Figure~\ref{fig:bidirectional_anchoring} contrasts the soft-label and hard-gate approaches.

The \texttt{power\_related} tag is used only for downstream analysis and does not control candidate entry into the pipeline. Section~\ref{subsec:ablation_soft_label} quantifies the cost of hard power-gating by comparing the soft-label approach with a hard-gate configuration.

\subsubsection{Fragment Filter and Unigram Path}\label{subsubsec:fragment_filter}

The anchor-driven bigram extraction step is intentionally permissive and therefore captures some grammatical fragments in addition to lexical items. For example, 咋个上个月 (zage shanggeyue, ``what about last month'') is a productive interrogative pattern rather than a fixed expression. We apply a four-condition fragment filter to remove such constructions while preserving genuine compounds. First, candidates shorter than two Chinese characters are discarded unless they appear in the whitelist of verified dialect expressions. Second, bigrams consisting entirely of dialect particle characters (着, 呢, 嘛, 噶, 哈, 了, 得, 过) are removed because they are not lexical items. Third, short-stem suffix constructions are removed: if a candidate ends with a dialect suffix (呢 (ne), 噶 (ga), 嘛 (ma), 哈 (ha), or 着 (zhao)) and the remaining stem is shorter than two characters, the candidate is discarded unless it appears in a whitelist of 19 verified dialect expressions, including 不来电呢 (bu laidian ne, ``no electricity coming,'' with dialect particle), 着电 (zhao dian, ``electrified'' or ``electric shock''), 停电哈 (tingdian ha, ``power outage,'' with dialect softening particle), and 线路老火 (xianlu laohuo, ``severe line fault''). Fourth, a jieba segmentation check is applied: if jieba segments the candidate into multiple tokens and at least one resulting token is a pure dialect particle, the candidate is classified as a fragment. This condition captures cases such as 拿着 (na zhe, ``holding''), where the segmentation 拿 + 着 reveals the aspect-marker role of 着 rather than its dialect-marker role. Overall, the fragment filter removes approximately 85\% of raw bigram candidates.

A parallel unigram extraction path captures individual dialect words that co-occur with power-domain seed terms in the same message. A unigram is retained if it satisfies four criteria: a minimum frequency of five occurrences, passing the fragment filter, a frequency below 1\% of total tokens to exclude excessively common function words, and the presence of at least one character from the dialect-specific character set $\mathcal{D}$. The character set $\mathcal{D}$ was iteratively refined across experiment runs. The final set includes uniquely dialectal characters and expressions (着, 呢, 嘛, 噶, 哈, 了, 得, 咋, \maochar{}, 木, 挨, 扎实 (zhashi, ``very/severe''), 鬼火 (guihuo, ``frustrated''), and 黑隆冬 (heilongdong, ``very dark'')), excludes characters common in standard Chinese power vocabulary (电 (dian, electricity), 炸 (zha, explode or burn out)), and removes ambiguous characters (老 (lao, old), 过 (guo, over/past)) after they allowed standard Chinese terms such as 老百姓 (laobaixing, ordinary people), 养老金 (yanglaojin, pension), and 莫过于 (moguoyu, nothing more than) to enter the candidate set.

\subsection{Stage 2: Semantic Classification}\label{subsec:stage2_semantic_classification}

Candidates that pass the fragment filter are scored using two complementary signals: semantic proximity to power-domain seeds and lexical fixedness inferred from tokenization behavior. The SBERT backbone, \texttt{paraphrase-multilingual-MiniLM-L12-v2} \citep{bib24}, produces 384-dimensional L2-normalized embeddings for multilingual semantic similarity. A FAISS IndexFlatIP index \citep{bib29} over the 65 power-domain seed embeddings supports exact cosine similarity search through inner products over L2-normalized vectors. For each candidate $c$, the semantic similarity score is defined as the maximum cosine similarity to any power-domain seed:

\begin{equation}
\mathrm{sim}(c)=\max_{s_j\in S_{\text{seed}}}\cos(v_c,v_{s_j}) .
\label{eq:semantic_similarity}
\end{equation}

Here, $v_c$ and $v_{s_j}$ are the SBERT embeddings of the candidate and seed, respectively. The wordhood score measures lexical fixedness by examining whether jieba treats the candidate as an atomic unit:

\begin{equation}
\mathrm{wordhood}(c)=
\max\!\left(0,\; 1-\frac{k-1}{|c|}\right)-0.30\cdot\frac{p}{k}.
\label{eq:wordhood}
\end{equation}

Here, $k$ is the number of tokens produced by \texttt{jieba.lcut}$(c)$, $|c|$ is the character length of the candidate, and $p$ is the number of resulting tokens that are pure dialect particles. The first term penalizes multi-token segmentations and therefore rewards candidates that resist fragmentation. The second term adds a penalty when the segmentation includes dialect particles, reflecting the observation that particle-containing multi-token outputs are usually grammatical constructions. The combined score is a weighted sum:

\begin{equation}
\mathrm{score}(c)=0.6\cdot\mathrm{sim}(c)+0.4\cdot\mathrm{wordhood}(c).
\label{eq:combined_score}
\end{equation}

The weights $\alpha = 0.6$ and $\beta = 0.4$ were determined empirically during development. SBERT similarity provides the stronger signal for power-domain relevance, but it can assign high scores to common vocabulary items. The wordhood score provides a weaker but complementary signal by penalizing such fragmented or particle-like outputs. Candidates with scores at or above the auto-accept threshold $\tau_{\text{high}} = 0.78$ enter the initial lexicon directly. Candidates with scores below the rejection threshold $\tau_{\text{low}} = 0.45$ are discarded. The remaining candidates form the uncertain pool $U$ for active learning.

\subsection{Stage 3: Active Learning with Twin Oracles}\label{subsec:stage3_active_learning}

\begin{figure}[t]
\centering
\includegraphics[width=\textwidth]{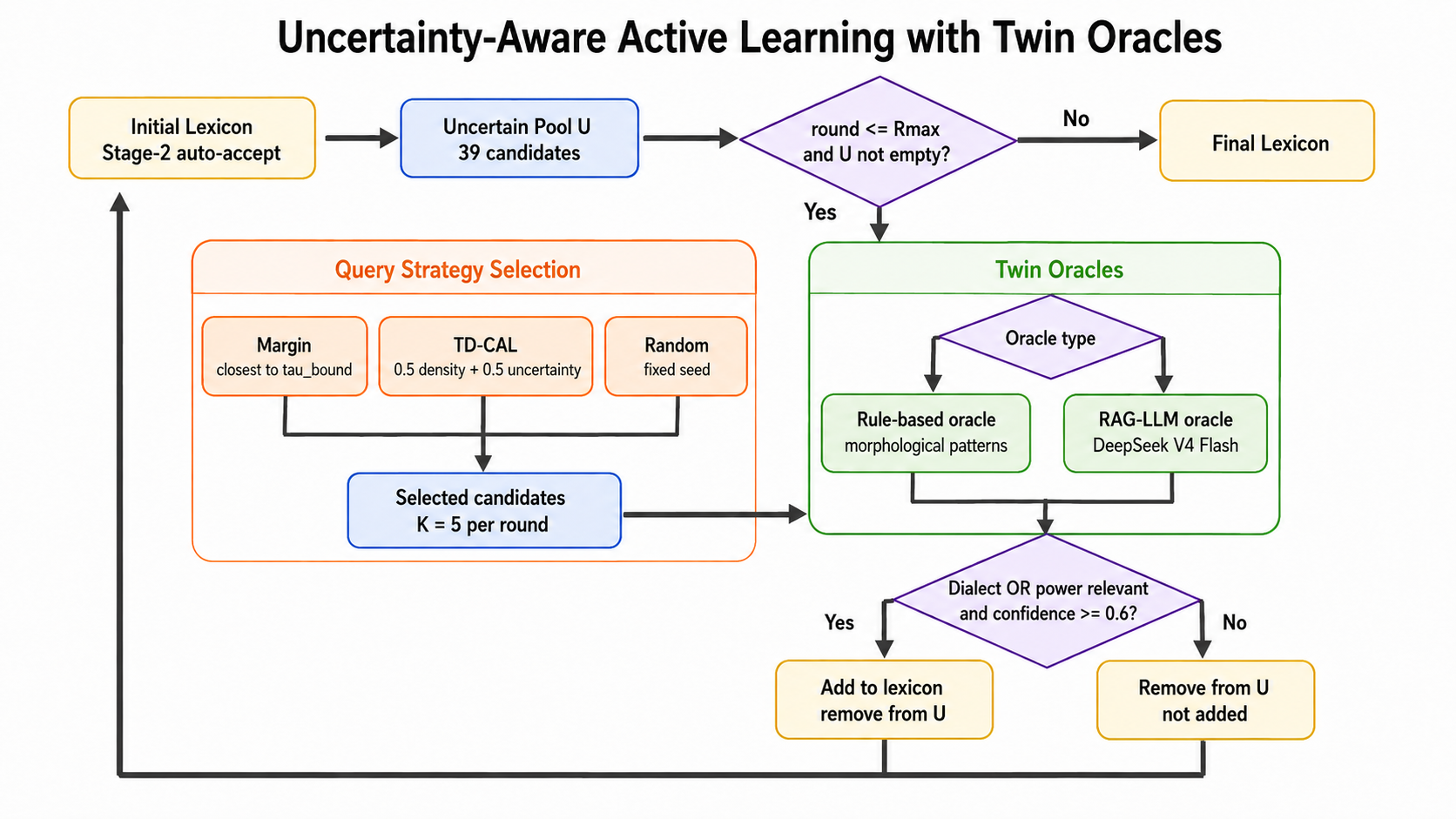}
\caption{Active learning loop: iterative query--verify--update cycle with three query strategies and twin oracles.}\label{fig:active_learning_loop}
\end{figure}

The uncertain pool $U$ from Stage 2 is passed to an iterative active learning loop, as shown in Figure~\ref{fig:active_learning_loop}. In each round, a query strategy selects $K$ candidates, the chosen oracle verifies them, accepted terms are added to the lexicon, and the loop continues for up to $R_{\text{max}}$ rounds.

The uncertain pool $U$ is the selection space for active learning. We compare three query strategies. Margin Sampling selects the top-$K$ candidates whose scores are closest to the decision boundary $\tau_{\text{bound}} = 0.60$, prioritizing candidates for which the classifier is least certain:

\begin{equation}
c_{\text{margin}}=
\mathop{\arg\min}_{c\in U}^{(K)}
|\mathrm{score}(c)-\tau_{\text{bound}}|.
\label{eq:margin_sampling}
\end{equation}

TD-CAL combines uncertainty with density. Uncertainty is computed from the combined score, and density is estimated using PMI. This choice is motivated by the limited usefulness of standard TF-IDF-based density for rare dialect pairs:

\begin{equation}
\mathrm{uncertainty}(c)=
1.0-\min\!\left(
\frac{|\mathrm{score}(c)-\tau_{\text{bound}}|}{m_{\max}},\,
1.0
\right),
\label{eq:uncertainty}
\end{equation}
where $m_{\max}=\max(\tau_{\text{high}}-\tau_{\text{bound}},\, \tau_{\text{bound}}-\tau_{\text{low}})=\max(0.18,0.15)=0.18$ denotes the maximum possible distance from the decision boundary within the uncertain score range $[\tau_{\text{low}},\tau_{\text{high}})$.

\begin{equation}
\mathrm{density}(c)=
\frac{\mathrm{PMI}(c)-\min_{c'\in U}\mathrm{PMI}(c')}
{\max_{c'\in U}\mathrm{PMI}(c')-\min_{c'\in U}\mathrm{PMI}(c')}.
\label{eq:density}
\end{equation}

\begin{equation}
\mathrm{score}_{\text{tdcal}}(c)=
0.5\cdot\mathrm{density}(c)+0.5\cdot\mathrm{uncertainty}(c).
\label{eq:tdcal_score}
\end{equation}

Random Sampling serves as a lower-bound baseline. It selects candidates uniformly from $U$ using a per-round seed (42 + round index), so that different random subsets are drawn in each round while experiment runs remain reproducible. We set $K = 5$ candidates per round and $R_{\text{max}} = 3$ rounds. This early stopping creates a regime where $|U| \gg K \times R_{\text{max}}$, allowing different query strategies to select different subsets. The oracle query count of 15 reported in Table~\ref{tab:main_results} corresponds to $K \times R_{\text{max}} = 5 \times 3$ queries per experiment. Auto-accepted candidates from Stage 2 enter the lexicon without consuming oracle queries in the default configuration.

Two oracle types are used to verify selected candidates. The \emph{rule-based oracle} applies dialect morphological pattern matching across three pattern categories: full-word matches, where the candidate contains any of 21 known Yuxi dialect expressions; suffix with power-character co-occurrence, where the candidate ends with a dialect suffix, the stem is at least 2 characters, and the word contains a power-domain character; and prefix matches, where the candidate begins with a dialect prefix and has sufficient length. The \emph{RAG-LLM oracle} retrieves the top-5 most relevant passages from a 30-entry bilingual power-domain knowledge base using SBERT embedding similarity, then constructs a two-dimensional assessment prompt for DeepSeek V4 Flash with temperature 0.0. The prompt assesses \texttt{has\_dialect\_features} and \texttt{power\_domain\_relevance} using 6 few-shot annotated examples. Acceptance requires $(\text{has\_dialect}\vee\text{power\_relevant})\wedge\text{confidence}\geq 0.6$. We use a permissive OR criterion at the oracle stage because the oracle is intended to retain potential dialect expressions, including borderline cases where the LLM may miss dialect features but identify power-domain relevance, or the reverse. The stricter AND criterion is applied only at the evaluation stage, where PDP measures the proportion of terms satisfying both conditions. This two-stage design reduces premature filtering of candidates that may be dialectal but lack explicit power-domain lexical markers, which is the type of candidate removed by hard power-gating. The low PDP values in Table~\ref{tab:main_results} (0.121--0.160) reflect this permissive design: many accepted terms are dialectal but not power-relevant, consistent with the soft-label strategy of deferring relevance judgments. Robust JSON extraction from LLM responses uses three fallback steps: standard \texttt{json.loads} with markdown code-block extraction, definition-field preprocessing to identify definition boundaries and escape unescaped quotes, and field-by-field regex extraction.

\subsection{Stage 4: Dual-Track LLM Evaluation}\label{subsec:stage4_llm_evaluation}

The final lexicon is evaluated by a separate LLM instance (DeepSeek V4 Pro) using the same two-dimensional assessment prompt but without acceptance gating. Using a different capability tier from the selection oracle (V4 Pro rather than V4 Flash) provides partial independence. The codebase also supports replacing the evaluator with a different model family, such as GLM-4-Plus or GPT-4o, for cross-model evaluation. Two quality metrics are defined:

\begin{equation}
\mathrm{DP}=
\frac{|\{w\in L:\mathrm{dialect}(w)=\mathrm{true}\}|}{|L|}.
\label{eq:dp}
\end{equation}

\begin{equation}
\mathrm{PDP}=
\frac{|\{w\in L:\mathrm{dialect}(w)=\mathrm{true}\wedge\mathrm{power}(w)=\mathrm{true}\}|}{|L|}.
\label{eq:pdp}
\end{equation}

DP (Dialect Precision) measures the proportion of lexicon terms that the evaluator identifies as carrying Yuxi dialect features. PDP (Power-Dialect Precision) applies a stricter criterion and measures the proportion of terms that are both dialectal and power-domain relevant. These metrics are calibrated against human expert annotations ($n=46$) via Cohen's $\kappa$, providing a reliability-checked basis for comparing experimental configurations.

\section{Experiments}\label{sec:experiments}

\subsection{Experimental Setup}\label{subsec:experimental_setup}

The experimental corpus contains 11,177 cleaned messages extracted from two WeChat group chat CSV exports covering approximately one year of Yuxi power-grid customer service interactions. After jieba tokenization with anchor and seed registration, the corpus contains 190,934 total tokens, 12,165 unique unigrams, and 48,395 unique bigrams. We also constructed a 30-entry bilingual knowledge base covering outage definitions, repair procedures, equipment descriptions, fault taxonomies, billing concepts, Yuxi dialect expressions with standard Chinese glosses, and morphological guidelines for distinguishing lexical items from grammatical patterns.

All experiments use a fixed random seed (42) for reproducibility. Table~\ref{tab:hyperparameters} summarizes the hyperparameter configuration. We compare six configurations: five soft-label configurations that vary the active learning query strategy (Margin, TD-CAL, Random) and oracle type (Rule, RAG-LLM), and one strict power-gating ablation (Margin + Rule). Candidate generation and semantic classification are kept constant across configurations, except for the power-gating condition in the ablation.

\begin{table}[t]
\caption{Hyperparameter configuration for all experiments.}\label{tab:hyperparameters}
\centering\footnotesize
\setlength{\tabcolsep}{3pt}
\begin{tabular}{p{0.27\columnwidth}p{0.28\columnwidth}p{0.34\columnwidth}}
\toprule
\textbf{Parameter} & \textbf{Value} & \textbf{Description} \\
\midrule
$\tau_{\text{high}}$ & 0.78 & Auto-accept threshold \\
$\tau_{\text{low}}$ & 0.45 & Rejection threshold \\
$\tau_{\text{bound}}$ & 0.60 & Decision boundary \\
$\alpha$ / $\beta$ & 0.6 / 0.4 & SBERT / wordhood weights \\
$K$ & 5 & Candidates per AL round \\
$R_{\text{max}}$ & 3 & Maximum AL iterations \\
PMI threshold & 1.5 & Minimum bigram association \\
SBERT model & \texttt{paraphrase-}\newline\texttt{multilingual-}\newline\texttt{MiniLM-L12-v2} & Embedding backbone \\
Oracle LLM & DeepSeek V4 Flash & Selection oracle (RAG-LLM) \\
Evaluator LLM & DeepSeek V4 Pro & Independent evaluator \\
LLM temperature & 0.0 & Deterministic decoding \\
LLM seed & 42 & Fixed seed for LLM \\
RAG top-$K$ & 5 & KB passages per query \\
Oracle min confidence & 0.6 & Minimum confidence \\
\botrule
\end{tabular}
\end{table}

\subsection{Candidate Generation and Filtering Performance}\label{subsec:candidate_generation_results}

The candidate generation pipeline extracted 349 candidates from the 11,177-message corpus, including 347 bigram candidates and 2 unigrams after tightening the dialect character set. The fragment filter removed 291 candidates (83.4\% of the total), leaving 58 candidates for semantic classification. The SBERT+wordhood classifier accepted 19 candidates into the initial lexicon (score $\geq 0.78$) and placed 39 candidates into the uncertain pool ($0.45 \leq$ score $< 0.78$); no candidates were rejected outright at this stage. Table~\ref{tab:main_results} reports the main results for all six configurations after the full pipeline, including the evaluator-derived decomposition into dialect ($|L_d|$) and power-dialect ($|L_{pd}|$) subsets.

\begin{table}[t]
\caption{Main experimental results for all six configurations. DP and PDP are computed by the independent LLM evaluator calibrated against $n=46$ human annotations. $|L_d|$ and $|L_{pd}|$ denote the sizes of the dialect and power-dialect lexicon subsets, respectively.}\label{tab:main_results}
\centering\scriptsize
\setlength{\tabcolsep}{1.6pt}
\renewcommand{\arraystretch}{1.08}
\begin{tabular}{p{0.34\columnwidth}cccccc}
\toprule
\textbf{Model} & \textbf{DP} & \textbf{PDP} & $\boldsymbol{|L_d|}$ & $\boldsymbol{|L_{pd}|}$ & \textbf{Lexicon} & \textbf{Queries} \\
\midrule
Margin + Rule (Soft-Label) & 1.0 & 0.160 & 25 & 4 & 25 & 15 \\
TD-CAL + Rule & 1.0 & 0.148 & 27 & 4 & 27 & 15 \\
Margin + LLM & 1.0 & 0.125 & 32 & 4 & 32 & 15 \\
Proposed (TD-CAL + LLM) & 1.0 & 0.125 & 32 & 4 & 32 & 15 \\
Random + LLM & 1.0 & 0.121 & 33 & 4 & 33 & 15 \\
Margin + Rule (Hard Gate) & 1.0 & 0.625 & 8 & \textbf{5} & 8 & 5 \\
\botrule
\end{tabular}
\end{table}

\begin{figure}[t]
\centering
\includegraphics[width=0.9\textwidth]{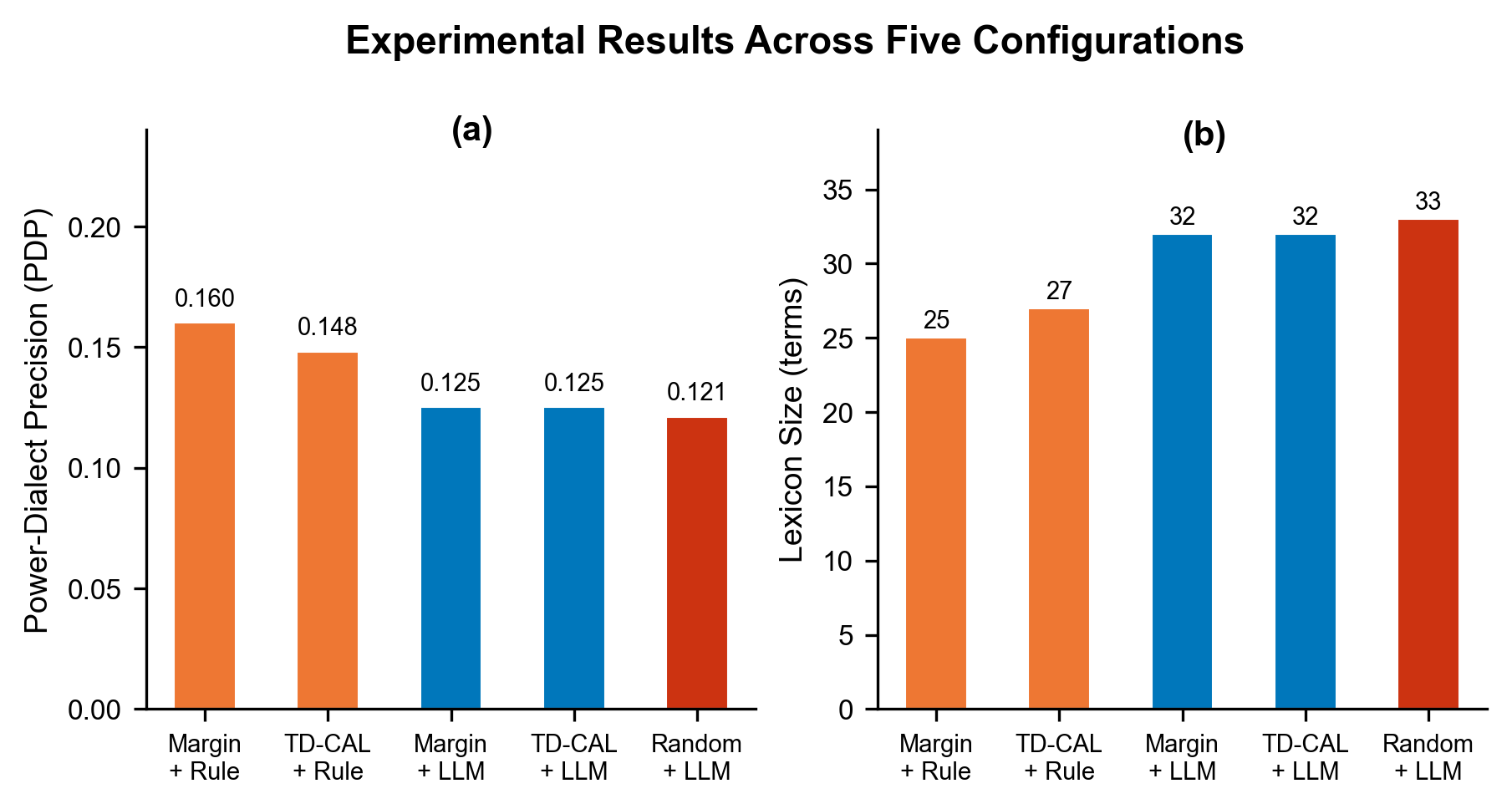}
\caption{(a) Power-Dialect Precision (PDP) and (b) lexicon size across the five experiment configurations. Rule-oracle configurations (left two bars) achieve higher PDP on smaller lexicons; LLM-oracle configurations (right three bars) produce larger lexicons with lower PDP. See Table~\ref{tab:main_results} for exact values.}\label{fig:main_results}
\end{figure}

Figure~\ref{fig:main_results} visualizes the trade-off between PDP and lexicon size across the five soft-label configurations.

Several patterns are observed. First, all six configurations achieve DP=1.0: every term in each constructed lexicon is identified by the independent evaluator as carrying Yuxi dialect features, with no standard Chinese false positives. This result is attributable to the combined effect of three precision-oriented mechanisms: (i) bidirectional anchoring restricts bigram candidates to expressions containing verified dialect markers; (ii) the fragment filter removes grammatical constructions while preserving lexical compounds; and (iii) the dialect character set $\mathcal{D}$ and jieba default-dictionary exclusion filter reduce non-dialect unigram entries. These components remove different types of noise and together support high dialect precision.

Oracle type also affects the final lexicons. The RAG-LLM oracle produces larger lexicons (32--33 terms) than the rule-based oracle (25--27 terms), a difference of approximately 24\%. This difference is consistent with the per-candidate acceptance rates: the LLM oracle accepts approximately 89\% of presented uncertain candidates, whereas the rule oracle accepts roughly 47\%. The higher LLM acceptance rate suggests that the LLM oracle recognizes dialect-related constructions beyond the rule oracle's morphological templates. In contrast, the rule oracle achieves higher PDP (0.148--0.160 vs.\ 0.121--0.125), indicating that its stricter morphological matching selects more strongly for power-domain relevance.

The query strategy comparison shows limited differentiation. With the LLM oracle, Margin, TD-CAL, and Random produce nearly identical lexicon sizes (32--33 terms) and the same $|L_{pd}|$ count (4). With the rule-based oracle, TD-CAL and Margin produce 27 and 25 terms, respectively, while both identify 4 power-dialect terms. This convergence, especially the uniform $|L_{pd}| = 4$ across all five soft-label configurations, suggests a ceiling effect: the corpus appears to contain only a small number of observable power-dialect bigrams under the current extraction criteria. The high LLM oracle acceptance rate ($\sim$89\%) further narrows the range in which query strategies can differ.

The power-dialect terms captured across configurations cover recurring complaint and query patterns in the Yuxi power-grid domain: 线路老火 (xianlu laohuo, literally ``line old fire,'' meaning ``severe line fault''), 不来电呢 (bu laidian ne, ``no electricity coming,'' with the dialect particle 呢 (ne)), 停电哈 (tingdian ha, ``power outage,'' with the dialect softening particle 哈 (ha)), 给有停电 (gei you tingdian, ``is there a power outage?,'' with the dialect question marker 给有 (gei you)), and 着电 (zhao dian, literally ``affected by electricity,'' meaning ``electric shock'' or ``electrified''). The evaluator classifies these terms as carrying both dialect features and power-domain relevance. The hard-gate configuration identifies one additional power-dialect term ($|L_{pd}| = 5$) from its smaller, more targeted candidate pool.

\subsection{Ablation Study: Soft-Label Versus Hard Power-Gating}\label{subsec:ablation_soft_label}

To isolate the contribution of soft-label layering, we compare two configurations that differ only in the power-gating strategy, using margin sampling with the rule-based oracle for three AL rounds. The hard-gate configuration applies the \texttt{power\_related} check as a binary filter: any bigram with $\mathrm{pr}(w_1,w_2) = 0$ is discarded before entering the pipeline. The soft-label configuration retains all dialect-anchored bigrams and stores the tag as metadata.

Table~\ref{tab:ablation} presents the ablation results across five pipeline stages. The largest reduction occurs at candidate extraction, where strict power gating reduces the raw candidate count from 349 to 41, removing 308 dialect bigrams (88.3\% of the total). These discarded bigrams include dialect conversational markers such as 咋个整 (zage zheng, ``what to do'') and 那咋整 (na zage zheng, ``then what to do''), which lack explicit power-domain lexical content but appear in outage complaint contexts as conversational signals associated with service requests. The reduction propagates through subsequent stages: the fragment-filter output decreases from 58 to 9 terms, the uncertain pool decreases from 39 to 5, and the final lexicon decreases from 25 to 8 terms. DP remains 1.0 in both conditions, suggesting that strict power gating does not improve dialect precision in this setting. The fragment filter, dialect character set, and jieba default-dictionary exclusion already remove major sources of non-dialect noise, whereas the strict gate removes additional dialect terms.

\begin{table}[t]
\caption{Soft-label versus strict power-gate ablation: pipeline throughput at each stage. Hard Gate applies the power-related check as a binary filter across all extraction paths (bigram, unigram, and anchor injection); Soft Label retains all dialect-anchored candidates. Both configurations use margin sampling with rule-based oracle at three AL rounds. $\Delta$ = percentage increase of Soft Label over Hard Gate.}\label{tab:ablation}
\centering\footnotesize
\setlength{\tabcolsep}{2.5pt}
\begin{tabular}{p{0.48\columnwidth}p{0.12\columnwidth}p{0.17\columnwidth}p{0.15\columnwidth}}
\toprule
\textbf{Pipeline Stage} & \textbf{Hard Gate} & \textbf{Soft Label} & $\boldsymbol{\Delta}$ \\
\midrule
Candidates extracted & 41 & 349 & +751\% \\
After fragment filter & 9 & 58 & +544\% \\
Uncertain pool $|U|$ & 5 & 39 & +680\% \\
Final lexicon $|L|$ & 8 & 25 & +213\% \\
DP & 1.0 & 1.0 & --- \\
\botrule
\end{tabular}
\end{table}

\begin{figure}[t]
\centering
\includegraphics[width=0.9\textwidth]{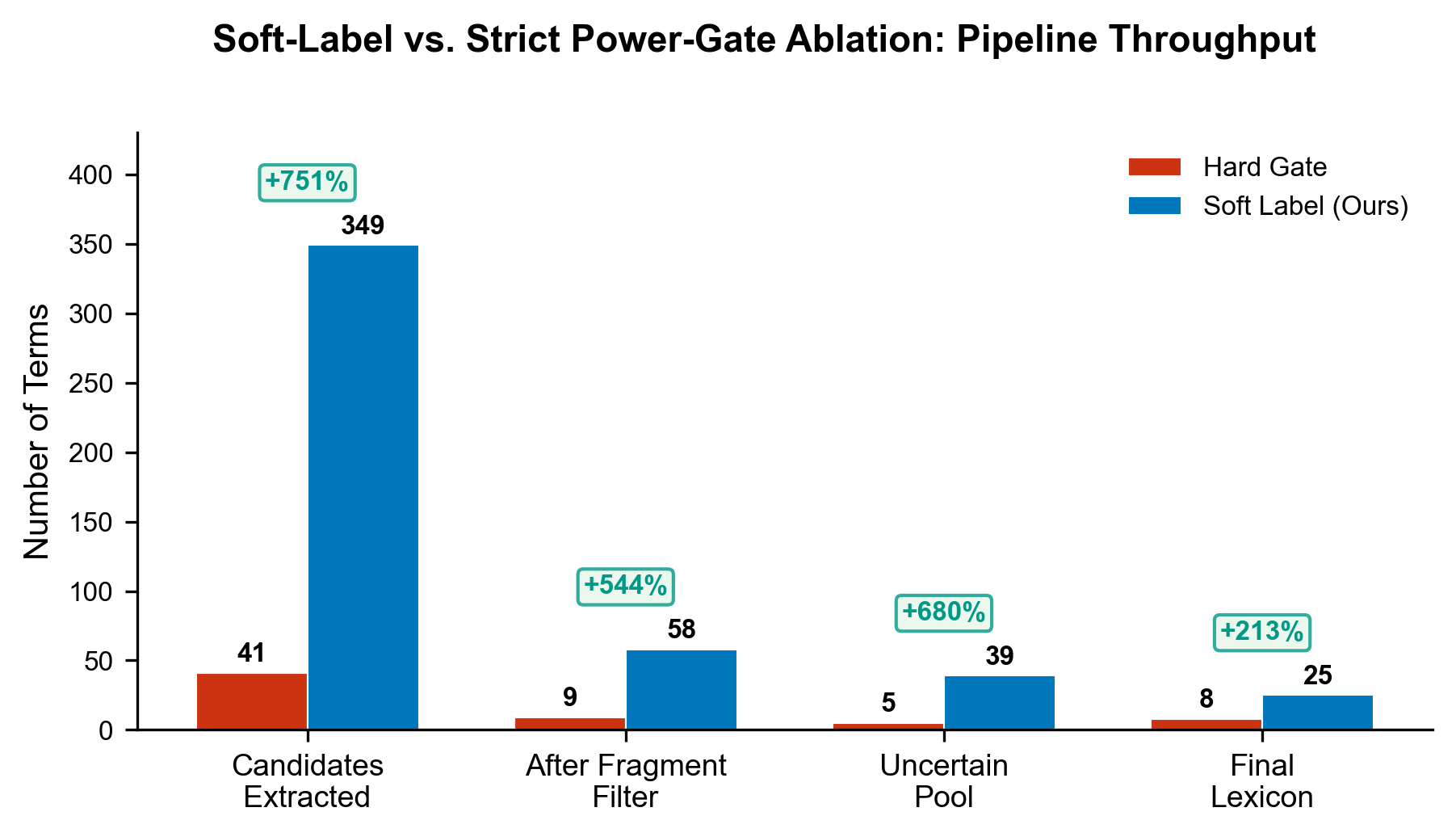}
\caption{Soft-label vs.\ hard gate ablation: pipeline throughput at each stage. The soft-label approach (blue) consistently retains more candidates, with the largest gain at the extraction stage (+751\%).}\label{fig:ablation}
\end{figure}

Figure~\ref{fig:ablation} visualizes the throughput comparison across pipeline stages.

The ablation suggests that relevance determination should be delayed until later stages, where more contextual and semantic information is available. Strict power gating makes an early filtering decision using character-level power-domain matching, thereby combining two distinct tasks: noise elimination, which is handled by the fragment filter, and relevance assessment, which is handled by the oracle and evaluator. By keeping these tasks separate, soft-label layering preserves candidate diversity while relying on downstream stages to maintain precision.

This comparison also shows an important trade-off. Although both approaches achieve DP=1.0, the hard gate yields higher PDP (0.625 vs.\ 0.160 for the comparable Margin+Rule soft-label configuration) and identifies more power-dialect terms ($|L_{pd}| = 5$ vs.\ 4), because it pre-filters for candidates with explicit power-domain lexical content. This comes at the cost of a 68\% reduction in total dialect terms relative to the soft-label configuration (8 vs.\ 25); equivalently, the soft-label configuration expands the final lexicon by 213\% over hard gating. The soft-label approach therefore trades power-domain precision for coverage, retaining dialect terms that may not be directly power-related. Whether this trade-off is useful depends on the downstream task, which remains to be evaluated.

\subsection{Cohen's Kappa Calibration}\label{subsec:kappa_calibration}

The LLM evaluator was calibrated against human expert annotations on $n = 46$ terms. These terms constitute the complete unduplicated lexicon pool obtained by merging all six experimental configurations (157 total entries reduced to 46 unique terms after deduplication). The calibration procedure was designed to select up to 100 terms, stratified into boundary-close ($|\text{combined score} - \text{decision boundary}| < 0.10$) and random subsets. Because the merged pool was smaller than the planned sample size, all 46 available terms were included (13 boundary-close and 33 random terms). A domain expert in Yuxi dialect and power-grid customer service assessed each term on the two retained evaluation dimensions. The \texttt{is\_fixed\_expression} dimension, originally included in the evaluation framework, was removed after an initial calibration round yielded $\kappa = 0.29$, indicating low agreement between the human and LLM annotators on this concept in dialect contexts.

Table~\ref{tab:kappa} presents the calibration results for the two retained dimensions. The \texttt{has\_dialect\_features} dimension shows perfect observed agreement: all 46 terms were annotated as dialectal by both the human expert and the LLM evaluator ($\kappa = 1.0$). This result should be interpreted cautiously because the label distribution is degenerate; when all samples share the same label, $\kappa$ provides limited discriminative information. For \texttt{power\_domain\_relevance}, the $\kappa$ value of 0.85 exceeds the conventional 0.6 threshold for substantial agreement and is above the 0.8 threshold often interpreted as near-perfect agreement \citep{bib35,bib58}. The confusion matrix contains 1 false negative, where the human annotator identified power relevance that the LLM evaluator missed, and 1 false positive, where the LLM evaluator identified power relevance that the human annotator did not. The remaining 44 of 46 terms were rated consistently.

\begin{table}[t]
\caption{Cohen's $\kappa$ between LLM evaluator (DeepSeek V4 Pro) and human expert annotations ($n=46$).}\label{tab:kappa}
\centering\footnotesize
\begin{tabular}{lccc}
\toprule
\textbf{Dimension} & $\boldsymbol{\kappa}$ & \textbf{95\% CI} & \textbf{Verdict} \\
\midrule
\texttt{has\_dialect\_features} & 1.0 & --- & Degenerate (all True) \\
\texttt{power\_domain\_relevance} & 0.85 & [0.64, 1.0] & Strong agreement \\
\botrule
\end{tabular}
\end{table}

The calibration sample of $n = 46$ supports the conclusion that $\kappa$ exceeds the 0.6 reliability threshold for the power-domain dimension. The improved $\kappa_{\text{power}} = 0.85$ (up from 0.75 in the initial $n=38$ calibration) indicates stable evaluator behavior on the expanded term set. The degenerate dialect result is inherent to the evaluated pool: a curated dialect lexicon is expected to contain dialect terms, so perfect agreement on this dimension is less informative than the power-domain relevance result.

\subsection{Pipeline Performance Evolution}\label{subsec:pipeline_evolution}

The pipeline evolved through three design iterations based on observed failure modes at different stages. Table~\ref{tab:pipeline_evolution} summarizes the quantitative progression.

\begin{table}[t]
\caption{Pipeline performance across three design iterations.}\label{tab:pipeline_evolution}
\centering\footnotesize
\begin{tabular}{lccccc}
\toprule
\textbf{Configuration} & \textbf{DP} & \textbf{PDP} & \textbf{Lexicon} & \textbf{Candidates} & \textbf{Unigram Noise} \\
\midrule
NPMI, no bidirectional anchoring & 0.17--0.28 & 0.01--0.05 & 101--148 & 347--496 & 72--75\% \\
PMI, strict power gate & 1.0 & 0.63 & 8 & 41 & 0\% \\
PMI, soft-label layering & 1.0 & 0.12--0.16 & 25--33 & 349 & 0\% \\
\botrule
\end{tabular}
\end{table}

The initial configuration used NPMI-based candidate extraction, a three-dimension evaluation framework, and no bidirectional anchoring mechanism. Although this configuration produced the largest lexicons (101--148 terms), unigrams accounted for 72--75\% of candidates and were dominated by standard Chinese vocabulary such as 供电所 (gongdiansuo, power supply office), 便民服务 (bianmin fuwu, convenience service), and 养老保险 (yanglao baoxian, pension insurance). These terms co-occurred with power-domain seed words in the corpus but did not carry dialect features, resulting in DP values of only 0.17--0.28 and PDP values near zero.

The introduction of PMI-based bidirectional anchoring as a hard power gate, together with a tighter dialect character filter and a two-dimension evaluation framework, eliminated unigram noise and achieved DP=1.0. However, the hard gate sharply reduced the candidate pool from several hundred terms to 41 and yielded lexicons of only 8 terms. This iteration improved precision but substantially reduced coverage.

The final configuration replaced the hard power gate with soft-label layering, used PMI as the TD-CAL density signal, and introduced early AL stopping at three rounds. This configuration expanded the candidate pool to 349 terms, an 8.5$\times$ increase over the hard-gate configuration, while maintaining DP=1.0. The larger uncertain pool (39 candidates vs.\ 5 under the hard gate) enabled multi-round active learning with strategy-specific selection patterns. Overall, the progression suggests that aggressive early filtering can remove useful signal along with noise, whereas deferring relevance decisions to later stages can preserve coverage while maintaining dialect precision.

\section{Discussion}\label{sec:discussion}

\subsection{Architectural Synergies}\label{subsec:architectural_synergies}

The high dialect precision observed across all configurations appears to result from a cascade of complementary design choices rather than from any single component. Bidirectional anchoring reduces non-dialect bigrams, the fragment filter removes grammatical constructions, and the tightened dialect character set blocks non-dialect unigrams. These components address different error sources, so their combined effect improves precision. At the same time, this design introduces a maintenance cost: each component must be recalibrated when the pipeline is transferred to a new dialect or domain.

The decision to use PMI rather than NPMI reflects an empirical trade-off specific to this corpus and pipeline. NPMI offers a useful theoretical advantage because normalization to $[-1, 1]$ supports cross-corpus threshold comparison. In our setting, however, two practical issues reduced its usefulness: the edit-distance-based typo filter removed legitimate multi-character compounds, including 线路老火 (xianlu laohuo, severe line fault), which was flagged as a typo of the seed term 线路 (xianlu, electrical line) at edit distance 2; and NPMI values for rare dialect pairs clustered tightly around 0.15--0.45, providing less discrimination than raw PMI values spanning 2.4--11.8. This result should not be interpreted as a general claim that PMI is superior to NPMI. Rather, it shows that downstream processing choices, such as the interaction between typo filtering and association thresholds, can affect the practical behavior of an otherwise well-motivated metric.

\subsection{Oracle Quality and the Limits of Active Learning Strategy Differentiation}\label{subsec:oracle_quality}

The reduced differentiation among query strategies under a high-acceptance oracle is consistent with prior observations that AL effectiveness depends on oracle characteristics \citep{bib30}. When an oracle accepts most presented candidates, the selection function of AL has limited room to affect the final lexicon. In our setting, the LLM oracle accepts 89\% of queried candidates, leaving only an 11\% rejection window in which strategy differences could appear. With only 39 uncertain candidates and a single random seed, this window is too narrow to support statistically distinguishable conclusions. The result should therefore be treated as suggestive and requires replication with larger candidate pools and multiple seeds.

\begin{figure}[t]
\centering
\includegraphics[width=0.9\textwidth]{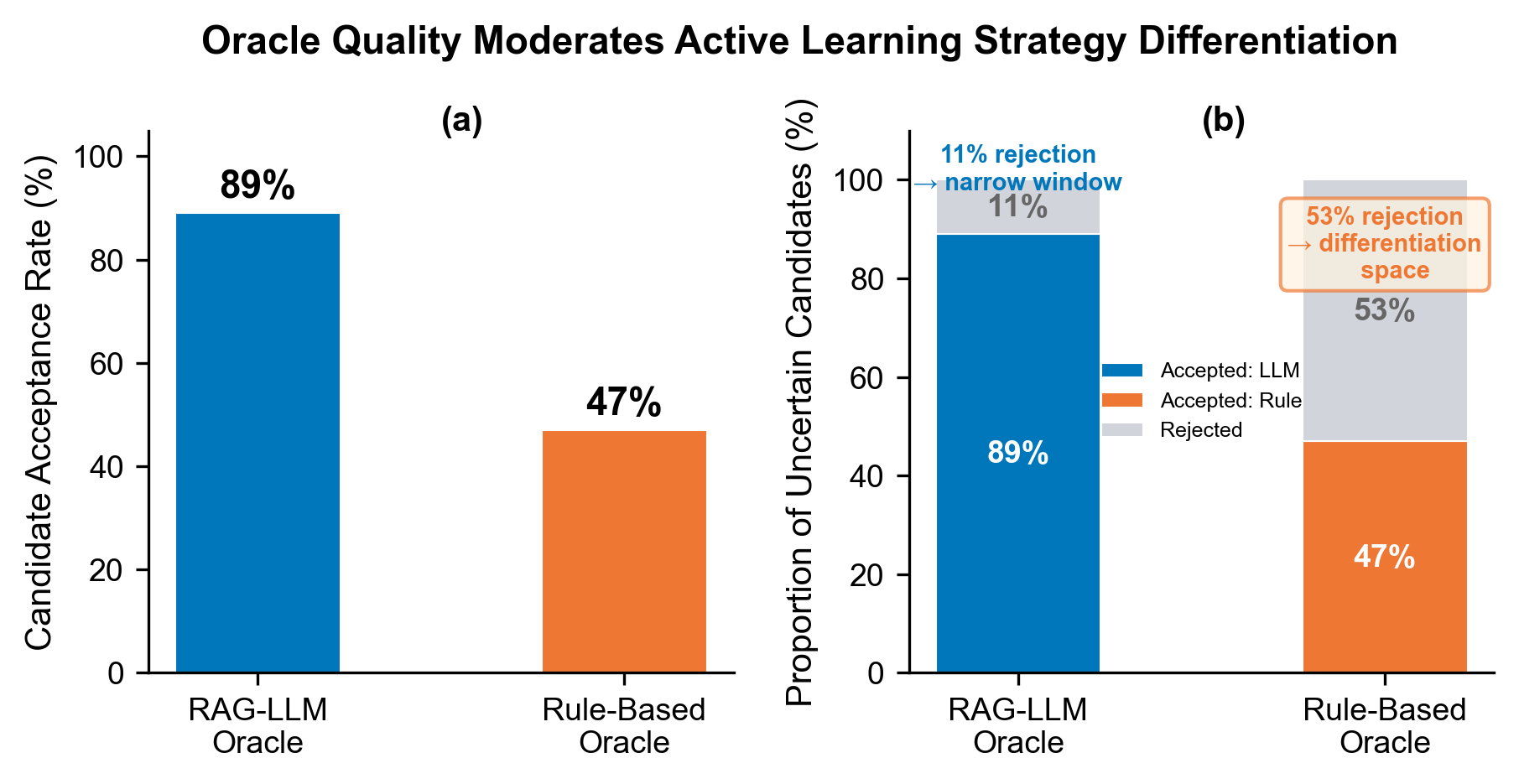}
\caption{(a) Oracle acceptance rates. (b) Stacked bar view: colored segments show accepted candidates where both oracles accept regardless of query strategy; gray segments show rejected candidates where strategies can differentiate (11\% for LLM vs.\ 53\% for Rule).}\label{fig:oracle_acceptance}
\end{figure}

Figure~\ref{fig:oracle_acceptance} illustrates this pattern. Panel (a) compares the raw acceptance rates, and panel (b) shows the resulting differentiation windows. The LLM oracle's 11\% rejection window leaves little room for strategy divergence, whereas the rule oracle's broader 53\% rejection window allows modest differentiation.

The rule-based oracle, with an acceptance rate of approximately 47\%, provides a wider discrimination window. Under this oracle, TD-CAL selects 27 terms and Margin selects 25, although both identify the same number of power-dialect terms ($|L_{pd}| = 4$). This pattern suggests a hypothesis for future work: the expected differentiation between AL strategies under oracle $O$ may depend on $\text{Var}[\text{accept}(O,c)]$ over the candidate distribution.

\subsection{Practical Significance and Limitations}\label{subsec:practical_significance}

The constructed lexicon contains 25--33 terms across configurations. Although modest in size, it includes high-frequency dialect expressions that are relevant to power-grid customer service interactions in the Yuxi region. These expressions encode distinctions related to fault severity, user emotional state, and request urgency, which may be useful for downstream customer service NLP systems. A domain expert could enumerate many of these terms manually; the value of the automated pipeline lies in its reproducibility across new data collections, its calibrated evaluation procedure, and its modular structure for adaptation to other dialects or service domains.

Several limitations constrain the generalizability of these findings. First, the evaluation is restricted to one dialect (Yuxi), one domain (power grid), and one platform (WeChat), and transfer to other settings has not been empirically tested. Second, the evaluation relies on precision-oriented metrics (DP and PDP) computed by an LLM evaluator. We do not report recall against an expert-constructed gold lexicon, nor do we provide full human evaluation of the final lexicons beyond the $n=46$ calibration set. As a result, we cannot quantify how much of the true regional power slang in the corpus is recovered. Third, the 11,177-message corpus limits the observable diversity of dialect bigrams, and additional power-dialect compounds may remain uncaptured. Fourth, the experimental comparisons are limited to internal variants of the proposed framework; comparisons with established phrase mining baselines such as AutoPhrase \citep{bib60}, SegPhrase \citep{bib61}, C-value/NC-value \citep{bib62}, TextRank \citep{bib63}, and zero-shot LLM extraction would provide stronger evidence for the contribution of individual components. Fifth, the active learning experiment is underpowered: with $|U| = 39$ uncertain candidates and a single random seed, strategy-level conclusions cannot be tested statistically. Sixth, the evaluator shares the same model family as the selection oracle, and cross-model evaluation has not yet been executed. Seventh, the lexicon has not been evaluated on downstream tasks such as complaint classification, urgency detection, or automated ticket routing. Finally, several pipeline choices, including the PMI threshold (1.5), SBERT/wordhood weights (0.6/0.4), anchor set composition (16 terms), seed set (65 terms), and whitelist (25 terms), were tuned through iterative experimentation on the same corpus without a held-out validation set; their sensitivity remains to be analyzed.

\subsection{Ethics and Reproducibility}\label{subsec:ethics_reproducibility}

The WeChat group data used in this study were collected from power-grid customer service groups operated by Yuxi Power Supply Bureau. Group members had consented to service-related message monitoring as part of the utility's customer engagement framework. Before analysis, personal identifiers, including \texttt{wxid\_} usernames, phone numbers, and profile names, were removed through regex-based anonymization. The resulting dataset cannot be publicly released because of privacy constraints and the terms of the original data-sharing agreement. Researchers seeking to replicate the study may request access through the corresponding author, subject to institutional data governance approval.

To support reproducibility despite data constraints, we provide the experiment code, hyperparameter configuration, anchor and seed sets, knowledge base prompts, oracle few-shot examples, and evaluation rubrics. The pipeline is version-controlled and produces deterministic outputs when the random seeds and corpus are fixed. Model artifacts, including the SBERT backbone, DeepSeek API versions, and configuration parameters, are documented in Table~\ref{tab:hyperparameters}. The calibration annotation template and the final constructed lexicons, including dialect and power-dialect subsets, are included in the experiment output directory.

\subsection{Future Directions}\label{subsec:future_directions}

Future work should address these limitations in several directions. Cross-dialect experiments, in which the Yuxi anchor set is replaced with markers from other regional Chinese dialects such as Sichuanese, Cantonese, or Northeastern Mandarin, would test the portability of the framework. An expert gold-standard lexicon constructed from a stratified corpus sample would enable recall-oriented evaluation through metrics such as recall, F1, and coverage. Comparisons with established phrase mining baselines, including AutoPhrase, SegPhrase, C-value/NC-value, TextRank, and zero-shot LLM extraction, would clarify the contribution of each pipeline component under a shared annotation protocol. Larger corpora, for example 100K+ messages per dialect, would improve coverage of long-tail dialect expressions and allow more reliable active learning comparisons with multiple random seeds and significance testing. Sensitivity analysis of the PMI threshold, score weights, anchor set, and seed set would quantify robustness. Further analysis of the relationship between oracle acceptance characteristics and AL strategy differentiation could test whether $\text{Var}[\text{accept}(O,c)]$ predicts strategy separation. Finally, downstream task evaluation and cross-model evaluator validation are needed to assess practical utility and reduce concerns about model-family circularity.

\section{Conclusion}\label{sec:conclusion}

This paper presented a statistical-anchor-driven weak-supervision framework for constructing a regional slang lexicon from noisy short texts in a low-resource dialect setting. The framework combines bidirectional anchoring with soft-label power-domain layering, allowing dialect-anchored candidates to be retained even when they do not contain explicit power-domain lexical markers. On 11,177 WeChat customer service messages from Yuxi, the framework reaches an LLM-evaluator-estimated dialect precision of DP=1.0. The ablation study shows that soft-label layering expands the candidate pool by 751\% and the final lexicon by 213\% relative to strict power-gating while maintaining the same dialect precision. The dual-track LLM evaluation framework, calibrated against human expert annotations ($\kappa_{\text{dialect}}=1.0$, $\kappa_{\text{power}}=0.85$, $n=46$), supports comparison across configurations, although cross-model validation and larger-scale human evaluation remain necessary.

The experiments also provide preliminary evidence that oracle characteristics affect active learning outcomes. Under the high-acceptance LLM oracle ($\sim$89\% acceptance), all three query strategies converge to the same $|L_{pd}|$, whereas modest differentiation appears only in total lexicon size under the more selective rule-based oracle ($\sim$47\% acceptance). This finding should be interpreted cautiously because the uncertain pool is small and only one random seed is used. Larger candidate pools, multiple seeds, and formal statistical testing are needed before drawing general conclusions about the interaction between oracle properties and AL efficiency.

The modular structure of the framework separates candidate generation, semantic classification, oracle verification, and independent evaluation into distinct stages, which may support adaptation to other dialects or service domains. More broadly, the results suggest that weak-supervision pipelines for low-resource dialect settings can benefit from delaying relevance decisions until later stages, where richer semantic and contextual information is available. The practical utility of the constructed lexicon still needs to be tested through downstream tasks such as complaint classification, urgency detection, and ticket routing, as well as through cross-dialect generalization experiments.

\backmatter

\bmhead{Acknowledgments}
The authors would like to thank the Editors and Reviewers for their review efforts and helpful feedback.

\section*{Declarations}

\begin{itemize}
\item \textbf{Funding:} Not applicable.
\item \textbf{Conflict of interest/Competing interests:} The authors declare no conflicts of interest. The funders had no role in the design of the study; in the collection, analyses, or interpretation of data; in the writing of the manuscript; or in the decision to publish the results.
\item \textbf{Ethics approval and consent to participate:} The WeChat group data used in this study were collected from power-grid customer service groups operated by Yuxi Power Supply Bureau. Group members had consented to service-related message monitoring as part of the utility's customer engagement framework.
\item \textbf{Data availability:} The dataset cannot be publicly released because of privacy constraints and the terms of the original data-sharing agreement. Researchers seeking to replicate the study may request access through the corresponding author, subject to institutional data governance approval.
\item \textbf{Code availability:} The complete source code, selected data and experimental scripts are publicly available at \url{https://github.com/zxzang/Construction-Method-of-Regional-Slang-Lexicon-for-Electricity-Demands}.
\item \textbf{Author contribution:} All authors contributed to the study conception and design. R.C. and L.S. collected the data and provided domain expertise. Z.Z. designed the methodology and experiments. All authors reviewed and approved the final manuscript.
\end{itemize}

\bibliography{sn-bibliography}

\end{CJK*}
\end{document}